\documentclass{article}

     \usepackage[final]{neurips_2021}

\usepackage[utf8]{inputenc} 
\usepackage[T1]{fontenc}    
\usepackage{hyperref}       
\usepackage{url}            
\usepackage{booktabs}       
\usepackage{amsfonts}       
\usepackage{nicefrac}       
\usepackage{microtype}      
\usepackage{xcolor}         
\usepackage{graphicx}
\usepackage{amsmath}
\usepackage{hhline}
\usepackage{adjustbox}

\title{As easy as APC: 
overcoming missing data and class imbalance in time series with self-supervised learning}

\author{
  Fiorella Wever  \\
  University of Amsterdam, Netherlands \\
  Clue by BioWink GmbH, Germany \\
  \texttt{fiorella.wever@gmail.com} \\
  \And
  T. Anderson Keller \\
  University of Amsterdam, Netherlands\\
  \texttt{t.anderson.keller@gmail.com} \\
  \And
  Laura Symul\\
  Department of Statistics, \\
  Stanford University, California \\
  \texttt{lsymul@stanford.edu} \\
  \And
  Victor Garcia\\
  University of Amsterdam, Netherlands \\
  \texttt{v.garciasatorras@uva.nl} \\
}

\begin{document}

\maketitle

\begin{abstract}
  High levels of missing data and strong class imbalance are ubiquitous challenges that are often presented simultaneously in real-world time series data. 
  Existing methods approach these problems separately, frequently making significant  assumptions about the underlying data generation process in order to lessen the impact of missing information. In this work, we instead demonstrate how a general self-supervised training method, namely Autoregressive Predictive Coding (APC), can be leveraged to overcome \textit{both} missing data and class imbalance simultaneously without strong assumptions. 
  Specifically, on a synthetic dataset, we show that standard baselines are substantially improved upon through the use of APC, yielding the greatest gains in the combined setting of high missingness and severe class imbalance. 
  We further apply APC on two real-world medical time-series datasets, and show that APC improves the classification performance in all settings, ultimately achieving state-of-the-art AUPRC results on the Physionet benchmark.  
\end{abstract}

\section{Introduction}
\looseness=-1
Rare event prediction on time series data is known to be a difficult challenge on its own  [2]. 
The difficulty often arises from the inherent class imbalance associated with rarity, encouraging models to classify most examples into the majority class, frequently resulting in poor performance for the minority class of interest [2].
Compounding this challenge is the ubiquity of missing data in real-world time series, further reducing the available information for classification on imbalanced datasets.
Prior work has demonstrated techniques for overcoming these obstacles, 
however most methods tackle each challenge separately; either dealing with properly learning in the context of missingness [4, 6, 18]
or improving identification of the minority class(es) [1, 5, 21, 22].

Recently, self-supervised pre-training has gained popularity, especially within the Natural Language Processing (NLP) community, showing significant performance improvements on downstream supervised learning tasks [3, 16, 17]. 
In this work, we propose leveraging a self-supervised autoregressive pre-training framework, namely Autoregressive Predictive Coding (APC) [9], to overcome both the challenges of class imbalance and missing data simultaneously in the context of time-series classification. 
Specifically, we suggest that pre-training recurrent neural network feature extractors to perform multiple-step-ahead forward prediction encourages the learning of discriminative representations which are independent of class labels while simultaneously taking advantage of potentially informative missingness patterns -- all without making strong assumptions about the data. 

We validate our approach on a synthetic dataset generated with varying levels of missingness and class imbalance, and find that APC substantially improves performance, especially in the severe class imbalance and high sparsity setting.
Further, we apply APC on two real-world datasets, Physionet and Clue, both containing missing values and class imbalance, and show that it improves the classification results in both cases, ultimately achieving  state-of-the-art AUPRC scores on the Physionet benchmark.

\section{Related work}
\paragraph{Class Imbalance}

Some of the most common methods for handling class imbalance consist of data re-sampling, such as under-sampling the majority class [1, 21]
and over-sampling the minority class [5].
However, under-sampling can result in losing a significant amount of data, 
while over-sampling methods, such as SMOTE [5],
artificially increase the data set.
Another widely adopted method is class weights [22],
which does not involve re-sampling, but provides an incentive to properly classify minority class samples by giving them more weights in the loss function. Most related to this work are methods that approach class imbalance through unsupervised learning. One such method is the dual autoencoders features (DAF) [14], which 
uses two stacked autoencoders to learn different sets of features of the data in an imbalanced setting. 
It is shown to outperform traditional data re-sampling methods, and suggests that leveraging unsupervised learning to obtain a set of relevant features could be a promising approach to deal with class imbalance. 
Our approach differs from those involving autoencoders in that a time shifting factor \textit{n} $\geq$ 1 encourages the  model to learn forward predictive features, and thus more global structure of the data, rather than only local structure [9]. 

\paragraph{Missing Data}
\looseness=-1
Traditional methods for handling missing data often involve first filling in the missing values, 
and then applying predictive models on the imputed data [6].
Choosing a suitable imputation scheme is complex, dataset-specific and relies on a good amount of domain expertise. Furthermore, this results in a two-step process that prevents
the prediction model from properly exploring the missingness patterns
[6].
As shown in [6], such \textit{informative missingness} may 
actually encode useful information about the target labels.
When it comes to state-of-the-art time series models, there are only a handful that incorporate these missingness patterns when learning from the data. One example is the Bidirectional Recurrent Imputation for Time Series (BRITS) [4],
which simultaneously imputes the missing values and performs classification/regression within a joint neural graph.
Another similar method is the GRU with trainable Decays (GRU-D) [6].
Both these methods
take advantage of two representations of informative missingness: \textit{masking} and \textit{time interval} [6].
Recently, two models based on ordinary differential equations, the ODE-RNN and the Latent-ODE, have also shown promising results on irregularly-sampled data [18].
However, the computational complexity of these models is high, which [12] 
concluded led to not finding the optimal hyperparameters.


\paragraph{Self-Supervised Pre-Training}
\looseness=-1
A special case of semi-supervised learning, called self-supervised pre-training, aims to learn a good initialization point for the supervised setting instead of changing the supervised learning objective [16].
Most models used for this purpose are based on autoencoders [19], 
while some of the most recent and promising methods 
are based on the idea of predictive coding, such as Contrastive Predictive Coding (CPC) [15]
and Autoregressive Predictive Coding (APC) [9].
Similar to CPC, APC learns from sequential data by trying to predict \textit{n} $\geq$ 1 steps ahead of the current step [9]. However, while CPC learns from forward prediction in a contrastive way, APC does so in an autoregressive manner, and has shown to significantly outperform CPC on speech recognition and phonetic classification tasks [8, 9]. Although APC pre-training has been previously proposed to improve supervised tasks, as far as we are concerned this is the first exploration of its application to handle both missing values and class imbalance. 

\section{Methods}
\paragraph{APC with MaskedMSE}

Given a length $N$ sequence $\{\mathbf{x}_t\}_{t=1}^N$ of observation vectors $\mathbf{x}_t \in \mathbb{R}^m$, the APC framework trains an autoregressive encoder $\mathrm{Enc}$ to sequentially reconstruct the observation at $n \geq 1$ time steps ahead of the current step, for $t=0$ to $N-n$ (see Figure \ref{APC}). Denoting $\mathbf{h}_{t}$ the hidden state of the encoder at time $t$, with $\mathbf{h}_0 = \mathbf{0}$, the output of the network $\mathbf{y}_t \in \mathbb{R}^{m}$ is given as:
\begin{equation}
\begin{array}{l}
\mathbf{h}_t = \mathrm{Enc}(\mathbf{h}_{t-1}, \mathbf{x}_t) \hspace{10mm}
\mathbf{y}_t = \mathbf{W}\mathbf{h}_t
\end{array}
\end{equation} 
\looseness=-1
In prior work [9],
the model was then trained to minimize the L1 distance between all predictions and ground truth (i.e $\mathcal{L} = \sum_{t=1}^{N-n}||\mathbf{x}_{t+n} - \mathbf{y}_t||_1$). However, if applied directly in the highly-missing data setting, we observe this loss encourages over-emphasis of missing predictions (analogous to the challenges encountered with class imbalance).
We thus propose a MaskedMSE loss to minimize reconstruction error over \emph{observed time steps} only. Formally:
\begin{equation}
\textbf{MaskedMSE}(\{\mathbf{x_t}, \mathbf{y}_t\}_{t=1}^N) = \frac{\sum^{\text{N-n}}_{t=1}(\mathbf{x}_{t+n} - \mathbf{y}_{t})^{2} \cdot \mathbf{m}_{t+n}}{\sum^{\text{N}-n}_{t=1}\mathbf{m}_{t+n}}
\end{equation} 
\noindent where
the masking vector $\mathbf{m}_{t+n} \in\{0,1\}^{m}$ indicates which variables are missing at time step $t+n$, and ensures that the encoder does not become biased towards missing values.

\paragraph{Encoder Models \& Missingness Handling}
To evaluate the impact of specific RNN encoder architectures within APC and how they handle missing values, we
compare two different encoders: 1) a baseline GRU
[7], 
and 2) the state-of-the-art GRU-D [6]
built to handle missingness. 
%
Since the GRU architecture does not inherently handle missing values, we add a binary ``missingness'' flag next to each measurement $x_{t}^{d}$, showing $1$ if the value is missing at that time-step and $0$ otherwise. 
GRU-D builds on the \textbf{GRU}, but with trainable \textbf{D}ecays and takes advantage of two representations of informative missingness patterns, \textit{masking} and \textit{time interval}, which replace the ``missingness'' flags of the GRU.
The \textit{time interval} $\delta_{t}^{d} \in \mathbb{R}$ indicates how long it has been for each variable \textit{d} since its last observation, and is used to exponentially \textit{decay} missing input variables and hidden states from their last value toward their empirical mean.
Representations learned in the self-supervised setting with APC are leveraged by using $\textbf{h}_t$, the output of the last layer of the APC encoder (GRU/GRU-D), as input for the classifier, trained with a cross-entropy loss.

\section{Experiments}

\paragraph{Datasets} We first evaluate our method on a synthetic dataset for which we control the level of missing values, as well as the level of class imbalance. This provides insights on the potential gains in performance provided by APC in handling class imbalance and data missingness both independently and in combination. 
The synthetic dataset consists of time-series of two variables, one continuous (noisy sine curve) and one binary (random Bernoulli). The output is a non-linear function of the sine curve frequency and of the Bernoulli variable probability (See Appendix for details).
We also experiment on two real-world datasets: a benchmark dataset from the Physionet Challenge 2012 [11], and time series data from the menstrual cycle tracking app Clue [10]. 
Clue samples contain 70.29$\%$ missing values on average with 92$\%$ majority class, while Physionet contains 79.75$\%$ missing with 86$\%$ majority class. 
Our code is available at \url{https://github.com/fiorella-wever/APC}.


\paragraph{Performance Evaluation}
For binary classification tasks (such as our synthetic dataset and Physionet), we choose to report AUPRC score as it more accurately shows how well the model is handling the minority class. 
For multi-class classification in the class imbalance setting (as with Clue), the weighted F1 score 
is the most common metric [20].
We also introduce the weighted F1 minority score (defined in appendix) to highlight the performances on the minority classes. 
Similar to prior work on Physionet [12], we use 60 $\%$ of the dataset as our training data, 20 $\%$ as validation and 20 $\%$ as the test set, preserving the class distribution in each set.
Evaluation on the test set was performed using the model with the best performance on the validation set.
Results are reported as the mean $\pm$ std. of the performances on the test set over 3 \textit{independent} runs.

\paragraph{Baselines}
For Physionet, we compare our results to the state-of-the art results for this benchmark dataset, achieved with GRU-D [12].
We note our reimplementation does not reach the same performance as the published model, and address the main differences between the implementations in the appendix. Despite these differences, our model still surpasses the best published AUPRC score for Physionet, as can be seen in Table 1. For Clue, given the proprietary nature of the dataset, we compare our results to a naive majority class baseline as well as ablations of our own model, observing APC augmented models again perform best.  
We additionally compare the performance of APC time shift factor \textit{n} $\geq$ 1 to an autoencoder (\textit{n} = 0), observing optimal performance with \textit{n} = 1, validating the use of APC compared to standard auto-encoding. As additional baselines, we compare our APC implementations to several different types of imputation methods (GRU-Mean, GRU-Forward, GRU-Simple) [12] 
in combination with class imbalance methods such as class weights.

\begin{figure}[ht]
\vskip 0.05in
\begin{center}
\centerline{\includegraphics[width=\columnwidth]{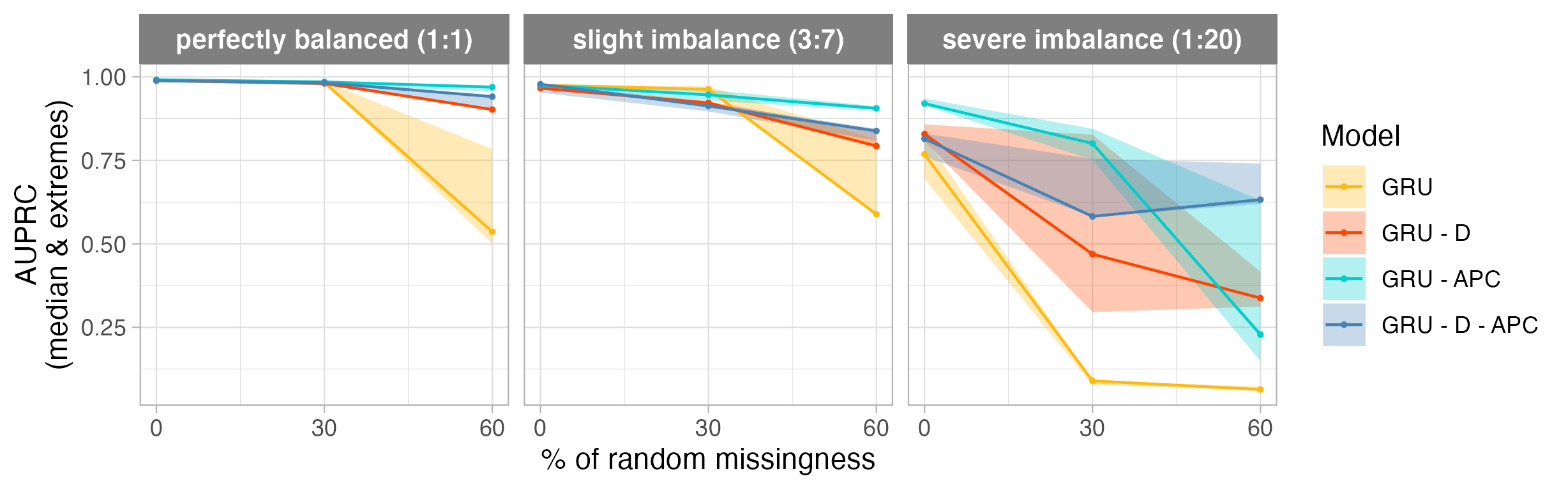}}
\caption{Classification results on synthetic dataset. GRU and GRU-D are implemented using class weights, while the APC implementations do not use class weights. The lines in the figure show the median, and the borders of the shaded region indicate the min and max (extremes) of the results.} 
\label{APC_sim_data}
\end{center}
\vskip -0.25in
\end{figure}

\paragraph{Results: Synthetic Data}
The results on our simulated dataset (Figure \ref{APC_sim_data}) show that applying APC improves the classification performance independently in the class imbalance setting and in the sparsity setting, as well as when dealing with both in conjunction.
Figure 1 highlights the median and min and max (extremes) of the results.
We see that APC improves performance substantially and significantly, especially in the high sparsity and severe class imbalance setting, improving the mean AUPRC score of the baseline GRU from $6.61\%$ to $66.42\%$ (appendix Table 6) when GRU-D and APC are applied in combination. Interestingly, we see that adding APC to a baseline GRU alone appears sufficient to achieve optimal performance, improving over GRU-D-APC in the majority of settings (see Table \ref{tab:results_synthetic}). As shown next, these results are further supported by the performance of APC models on real-world data, ultimately lending strength to the idea that self-supervised learning is a feasible solution to both class imbalance and data missingness \emph{on its own} without the need for data-dependent assumptions.

\paragraph{Results: Real-World Data}
Applied to real-world data, we find that APC with time shift factor \textit{n} = 1 
performs best on both Physionet (Table \ref{physio_results}) and Clue (appendix Tables \ref{tab:results_clue_APC_GRU} \& \ref{tab:results_clue_APC_GRUD}), outperforming standard autoencoding (\textit{n} = 0),
indicating that the features learned through forward prediction are more relevant than those learned by reconstructing the input directly. 
In Table \ref{physio_results}, we show that the GRU-APC, which tackles both class imbalance and missing data simultaneously, outperforms baselines that use a combination of independent methods.
We include four literature baselines to showcase the variability in existing results, and observe that our models leveraging APC outperform all of them, achieving state-of-the-art AUPRC scores and competitive AUROC scores without needing class weights.
Further, in Table \ref{clue_results}, we show a continuation of this trend, observing that GRU-APC also improves the results on Clue compared to baselines, although in this case oversampling + class weights are also needed likely due to the more severe imbalance.
Interestingly, we see a different trend in performance of the imputation methods on both datasets, 
indicating that these methods impose assumptions that are dataset-specific. 
One major advantage of APC is that it does not impose such strong assumptions on what the missing values should be, yet consistently achieves competitive performance.

\begin{table}[h!]
\parbox{.5\linewidth}{
\centering
\caption{Comparisons on Physionet 2012 mortality task. 
OS = oversampling, CW = class weights. \label{physio_results}}
\resizebox{0.5\columnwidth}{!}{\begin{tabular}{lccc}
\toprule
Model & Class imbalance & AUROC & AUPRC \\
& method & & \\
\midrule
GRU-D [6] & OS & 84.24 $\pm$ 1.2 & -\\
\hline
GRU-D [12] & OS & \textbf{86.3} $\pm$ \textbf{0.3} & 53.7 $\pm$ 0.9 \\
\hline 
GRU-D  [4] & none & 83.4 $\pm$ 0.2 & -\\
\hline 
BRITS [4] & none & 85.0 $\pm$ 0.2 & -\\
\hhline{====}
GRU & CW & 85.4 $\pm$ 0.4 & 51.4 $\pm$ 0.9\\ 
\hline 
GRU-Mean & CW & 84.6 $\pm$ 0.2 & 50.3 $\pm$ 0.7\\ 
\hline 
GRU-Forward & CW & 84.3 $\pm$ 1.0 & 52.0 $\pm$ 1.4\\ 
\hline 
GRU-Simple & CW & 85.5 $\pm$ 0.1 & 53.8 $\pm$ 0.2\\ 
\hline 
GRU-D & CW & 85.5 $\pm$ 0.3 & 53.1 $\pm$ 0.4\\
\hline 
GRU-APC (\textit{n} = 0) & CW & 84.2 $\pm$ 0.3 &  50.4 $\pm$ 0.3 \\
\hline 
GRU-D - APC (\textit{n} = 0) & CW & 85.0 $\pm$ 0.3 & 53.3 $\pm$ 0.3 \\
\hline 
GRU-APC (\textit{n} = 1) & none & \textbf{86.0} $\pm$ \textbf{0.5} & \textbf{54.1} $\pm$ \textbf{1.0}\\
\hline 
GRU-D - APC (\textit{n} = 1) & none & 85.2 $\pm$ 0.9  & 54.1 $\pm$ 2.3\\
\hline 
GRU-APC (\textit{n} = 1) & CW & 85.9 $\pm$ 0.3 & 53.5 $\pm$ 0.5\\ 
\hline 
GRU-D - APC (\textit{n} = 1) & CW &  85.3 $\pm$ 0.1 & \textbf{55.1} $\pm$ \textbf{0.9}\\ 
\bottomrule
\end{tabular}%
}
}
\hfill
\parbox{.4\linewidth}{
\centering
\caption{Comparisons on Clue classification task. \label{clue_results}}
\resizebox{0.4\columnwidth}{!}{
\begin{tabular}{lcc}
\toprule
Model  & weighted  & weighted  \\
 & F1 & F1 minority\\
\midrule
naive classifier &  87.47 & -\\
\hline 
GRU & 90.3 $\pm$ 0.7 & 22.3 $\pm$ 7.2\\ 
\hline 
GRU-Mean & 88.4 $\pm$ 0.5 & 22.1 $\pm$ 1.8\\ 
\hline 
GRU-Forward &  87.5 $\pm$ 0.2 & 22.1 $\pm$ 0.5 \\ 
\hline 
GRU-Simple & 87.8 $\pm$ 0.3 & 22.2 $\pm$ 0.4\\ 
\hline 
GRU-D & 87.5 $\pm$ 0.3 & 22.5 $\pm$ 0.7 \\
\hline 
GRU-APC (\textit{n} = 1) & \textbf{90.7} $\pm$ \textbf{0.1} & \textbf{25.7} $\pm$ \textbf{1.1}\\
\hline 
GRU-D - APC (\textit{n} = 1) & 90.3 $\pm$ 0.0 & \textbf{27.3} $\pm$ \textbf{0.5} \\
\bottomrule
\end{tabular}}
}
\end{table}

\section{Discussion}
In this work we demonstrate self-supervised APC pre-training as a promising method to deal with both severe class imbalance and high levels of missing data in the context of time-series classification.
Although APC pre-training is not a drastically novel method, the proposed application to handle missing values and class imbalance has not been significantly explored before, making our contribution a grounded exploration of this idea.
We show empirically on synthetic and real world data that representations learned by APC pre-training are superior to standard auto-encoding baselines, and additionally perform on-par or better than existing methods built explicitly to handle the challenges of imbalance and missingness. We note that this work is inherently preliminary and limited in scope, however we believe it provides a grounded foundation for further exploration of the benefits of self-supervised learning in the context of missing data and class imbalance.

\section*{References}


{
\small

[1] Barandela, R., Sánchez, J. S., Garcıa, V., \& Rangel, E. (2003). Strategies for learning in class imbalance problems. Pattern Recognition, 36(3), 849-851.

[2] Branco, P., Torgo, L., \& Ribeiro, R. (2015). A survey of predictive modelling under imbalanced distributions. arXiv preprint arXiv:1505.01658.

[3] Brown, T. B., Mann, B., Ryder, N., Subbiah, M., Kaplan, J., Dhariwal, P., ... \& Amodei, D. (2020). Language models are few-shot learners. arXiv preprint arXiv:2005.14165.

[4] Cao, W., Wang, D., Li, J., Zhou, H., Li, L., \& Li, Y. (2018). Brits: Bidirectional recurrent imputation for time series. arXiv preprint arXiv:1805.10572.

[5] Chawla, N. V., Bowyer, K. W., Hall, L. O., \& Kegelmeyer, W. P. (2002). SMOTE: synthetic minority over-sampling technique. Journal of artificial intelligence research, 16, 321-357.

[6] Che, Z., Purushotham, S., Cho, K., Sontag, D., \& Liu, Y. (2018). Recurrent neural networks for multivariate time series with missing values. Scientific reports, 8(1), 1-12.

[7] Cho, K., Van Merriënboer, B., Gulcehre, C., Bahdanau, D., Bougares, F., Schwenk, H., \& Bengio, Y. (2014). Learning phrase representations using RNN encoder-decoder for statistical machine translation. arXiv preprint arXiv:1406.1078.

[8] Chung, Y. A., \& Glass, J. (2020, May). Generative pre-training for speech with autoregressive predictive coding. In ICASSP 2020-2020 IEEE International Conference on Acoustics, Speech and Signal Processing (ICASSP) (pp. 3497-3501). IEEE.

[9] Chung, Y. A., Hsu, W. N., Tang, H., \& Glass, J. (2019). An unsupervised autoregressive model for speech representation learning. arXiv preprint arXiv:1904.03240.

[10] Clue   by   BioWink   GmbH   (2020).URLhttps://helloclue.com/. Adalbertstraße 7-8, 10999 Berlin,Germany. 

[11] Goldberger, A. L., Amaral, L. A., Glass, L., Hausdorff, J. M., Ivanov, P. C., Mark, R. G., ... \& Stanley, H. E. (2000). PhysioBank, PhysioToolkit, and PhysioNet: components of a new research resource for complex physiologic signals. circulation, 101(23), e215-e220.

[12] Horn, M., Moor, M., Bock, C., Rieck, B., \& Borgwardt, K. (2020, November). Set functions for time series. In International Conference on Machine Learning (pp. 4353-4363). PMLR.

[13] Mikalsen, K. Ø., Soguero-Ruiz, C., Bianchi, F. M., Revhaug, A., \& Jenssen, R. (2021). Time series cluster kernels to exploit informative missingness and incomplete label information. Pattern Recognition, 115, 107896.

[14] Ng, W. W., Zeng, G., Zhang, J., Yeung, D. S., \& Pedrycz, W. (2016). Dual autoencoders features for imbalance classification problem. Pattern Recognition, 60, 875-889.

[15] Oord, A. V. D., Li, Y., \& Vinyals, O. (2018). Representation learning with contrastive predictive coding. arXiv preprint arXiv:1807.03748.

[16] Radford, A., Narasimhan, K., Salimans, T., \& Sutskever, I. (2018). Improving language understanding by generative pre-training.

[17] Radford, A., Wu, J., Child, R., Luan, D., Amodei, D., \& Sutskever, I. (2019). Language models are unsupervised multitask learners. OpenAI blog, 1(8), 9.

[18] Rubanova, Y., Chen, R. T., \& Duvenaud, D. (2019). Latent odes for irregularly-sampled time series. arXiv preprint arXiv:1907.03907.

[19] Sagheer, A., \& Kotb, M. (2019). Unsupervised pre-training of a deep LSTM-based stacked autoencoder for multivariate time series forecasting problems. Scientific reports, 9(1), 1-16.

[20] Sokolova, M., \& Lapalme, G. (2009). A systematic analysis of performance measures for classification tasks. Information processing \& management, 45(4), 427-437.

[21] Yen, S. J., \& Lee, Y. S. (2009). Cluster-based under-sampling approaches for imbalanced data distributions. Expert Systems with Applications, 36(3), 5718-5727.

[22] Zhu, M., Xia, J., Jin, X., Yan, M., Cai, G., Yan, J., \& Ning, G. (2018). Class weights random forest algorithm for processing class imbalanced medical data. IEEE Access, 6, 4641-4652.

}




\newpage
\appendix

\section{Appendix}
\subsection{\large Datasets}
\subsubsection*{Synthetic dataset}

The synthetic dataset is generated as $N$ two-variable time series of length $T$. Here $N = 2000$ and $T = 100$. The first variable, $x$, is a continuous variable simulated as $$x_n(t) = 1 + o_n + \cos(\frac{2\pi t}{P_n}) + \epsilon$$ with $\epsilon \sim  \mathcal{N}(0, \sigma) $ and where $o_n$, a random offset, is uniformly sampled from $[0,1]$ and $P_n$, the cosine period, is uniformly sampled from $[P_{min},P_{max}]$. Here, we chose $P_{min} = 5$ and $P_{max} = 20$. The second variable, $b$, is simulated as a Bernoulli variable with a probability $p_n$, which is randomly sampled in $[0,1]$ for each time series $n$.

The output $y_n$ is built as a combination of $x$'s period ($P_n$) and $b$'s probability ($p_n$). Specifically, $y_n = (P_n \leq 0.5) (p_n \geq 0.5)$.

\subsubsection*{Physionet}
For the benchmark dataset from the Physionet Challenge 2012 [11], the classification task is to predict whether an ICU patient will survive or die during their stay at the hospital.
Physionet samples contain 79.75 $\%$ missing values on average $[62.61, 99.94]$. Physionet is a \textit{binary} classification benchmark, where the majority class makes up 86$\%$ of all the samples.

\subsubsection*{Clue}
For the Clue dataset, we seek to predict the discontinuation of birth control methods over time.  
Clue samples contain 70.29 $\%$ missing values on average $[6.06, 81.45]$.
Since we're dealing with 4 different output classes (see Table \ref{tab:clue_class_imbalance}), Clue requires \textit{multi class} classification, where the majority class makes up 92 $\%$ of all the samples.

\subsection{\large Class imbalance}
\begin{table}[h]
\caption{\label{tab:physio_class_imbalance} Physionet: Class distribution.}
\vskip 0.15in
\begin{center}
\begin{sc}
\begin{tabular}{lc}
\toprule
Output Label & Percentage \% of total \\
\midrule
Survivor & 85.76 \\
Died in-hospital & 14.24 \\
\bottomrule
\end{tabular}
\end{sc}
\end{center}
\vskip -0.3in
\end{table}

\vspace{1cm}

\begin{table}[h]
\caption{\label{tab:clue_class_imbalance} Clue: Class distribution}
\vskip 0.15in
\begin{center}
\begin{sc}
\begin{tabular}{lc}
\toprule
Output Label & Percentage \% of total \\
\midrule
ON &  91.52 \\
OFF &  4.88 \\
OTHER-hormonal & 2.22 \\
OTHER-non-hormonal & 1.38 \\
\bottomrule
\end{tabular}
\end{sc}
\end{center}
\vskip -0.3in
\end{table} 

\newpage
\subsection{\large Missing values}
Clue contains 70.29 $\%$ missing values on average ranging from 6.06 $\%$ to 81.45 $\%$, while Physionet contains 79.75 $\%$ missing values on average ranging from 62.61 $\%$ to 99.94 $\%$. 

\begin{figure}[h]
\vskip 0.2in
\begin{center}
\centerline{\includegraphics[width=\columnwidth]{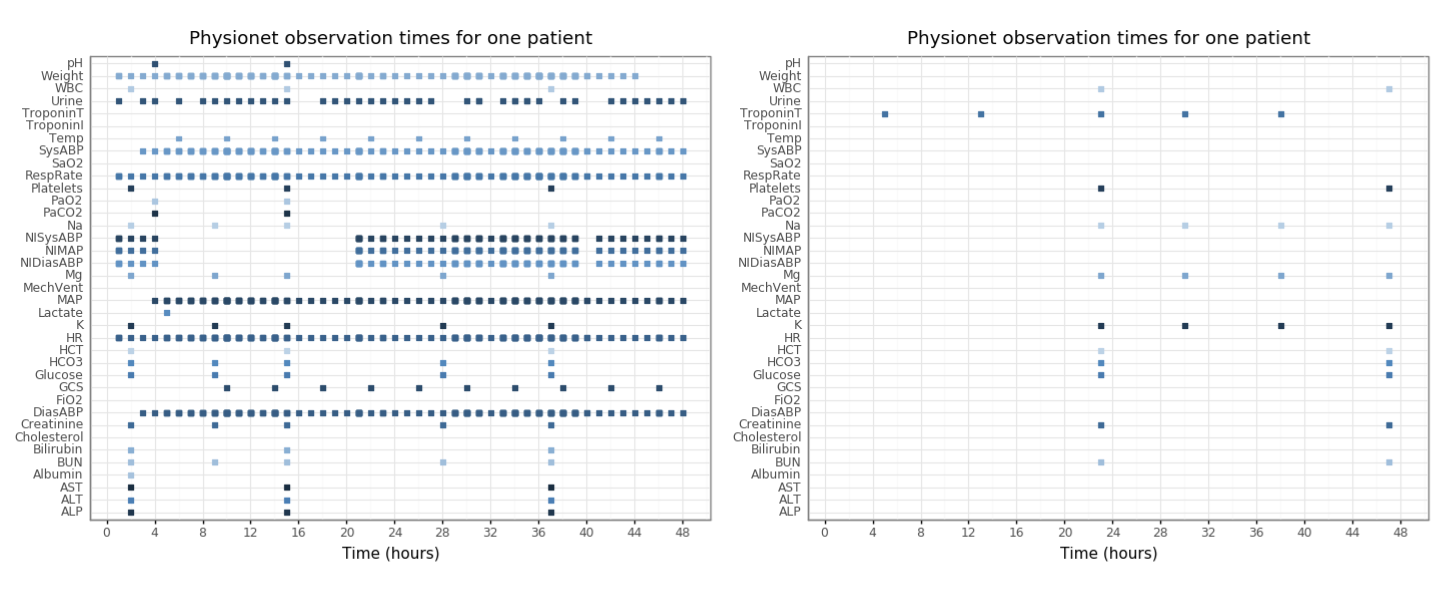}}
\caption{Physionet: Examples of time series for two different ICU patients with different frequencies of observations. On the left, we have a time series with a high frequency of observations. On the right is an example of a patient who is barely observed.} 
\label{physio_high_low_missing}
\end{center}
\end{figure}

\begin{figure}[h]
\vskip 0.2in
\begin{center}
\centerline{\includegraphics[width=\columnwidth]{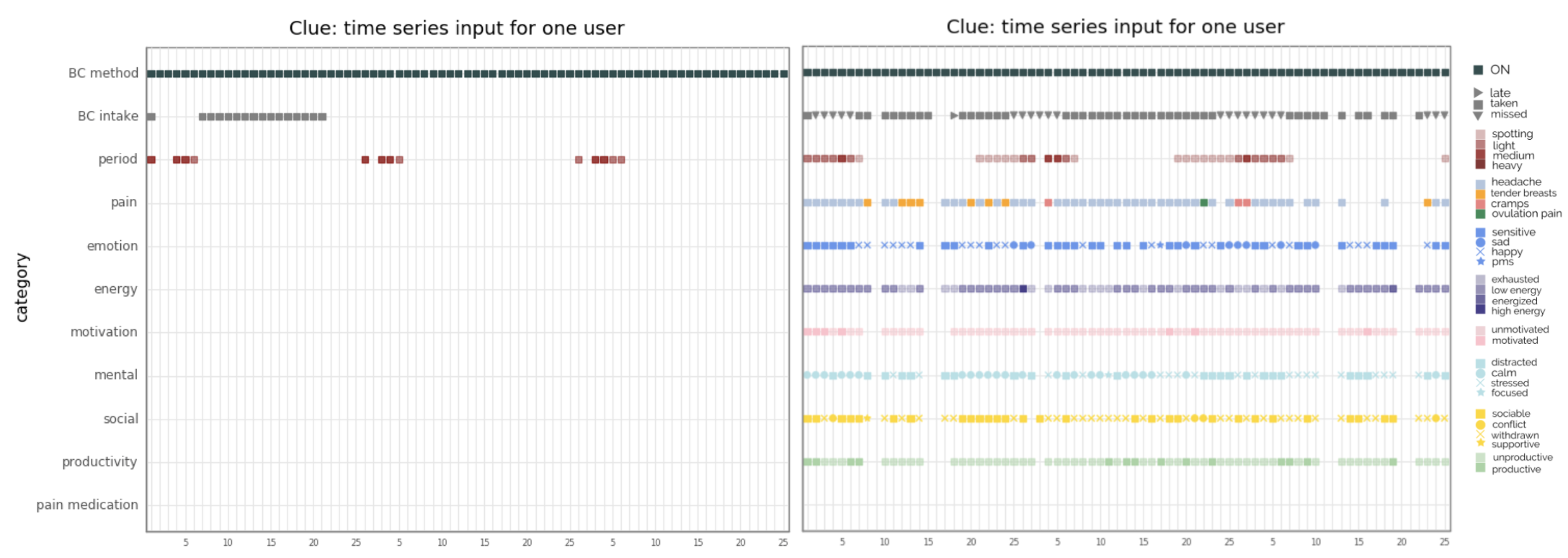}}
\caption{Clue: Example of the time series of the 3 input cycles for a user that tracks with low frequency (left), and a user that tracks with high frequency (right).} 
\label{clue_high_low_missing}
\end{center}
\end{figure}

\newpage
\subsection{\large Clue's Empirical Mean}
While Physionet has one mean per variable, we implement a multi-dimensional mean for the Clue dataset, with each variable having a different mean for each day of the cycle.

\subsection{\large Time scales}
In the Physionet dataset, the measurements are observed at irregular time steps
at a minute-by-minute resolution. For the GRU model, we aggregate the measurements to a 1 hour resolution, 
similar to what had been done in previous work on this dataset [6].
However, for the GRU-D model we aggregate the measurements to a 1-decimal time resolution (e.g. 2.6 hours) as in the original GRU-D paper [6].
This allows the model to learn with more detail the time interval since a variable was last observed.

\subsection{\large APC model architecture}
\begin{figure}[ht]
\vskip 0.05in
\begin{center}
\centerline{\includegraphics[width=0.6\columnwidth]{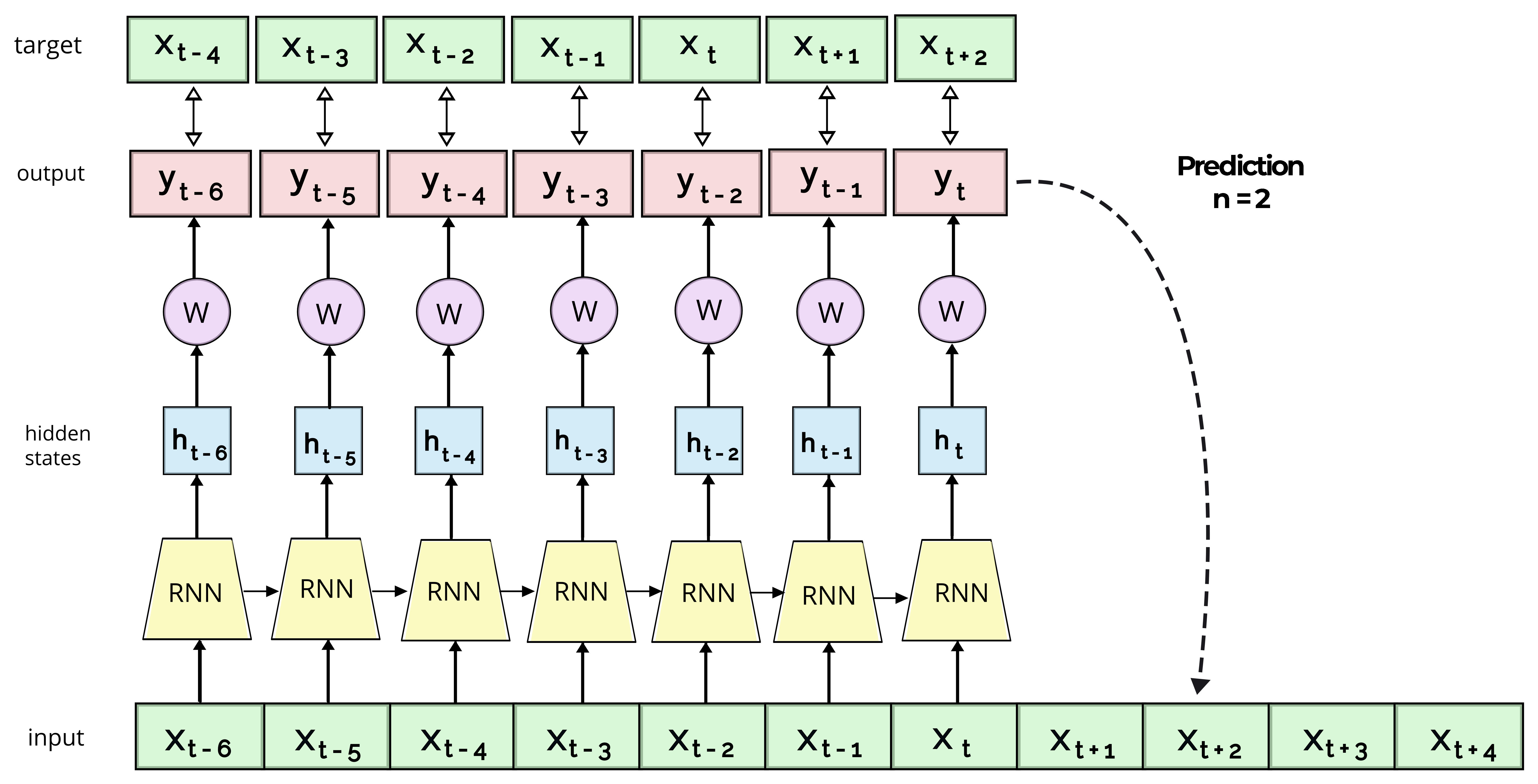}}
\caption{\textbf{Autoregressive Predictive Coding (APC)}: input sequence $\{\mathbf{x}_t\}_{t=1}^N$ is encoded (e.g. with an RNN) at each time step to a hidden state $\mathbf{h}_{t}$. A matrix $\mathbf{W}$ then transforms the hidden states to an output sequence $\{\mathbf{y}\}_{t=1}^N$ of the same dimension as $\mathbf{x}_t$.} 
\label{APC}
\end{center}
\vskip -0.25in
\end{figure}

\subsection{\large Frozen vs fine-tuned APC weights}
To leverage the representations that were learned in the self-supervised setting with APC, we take the output of the last layer of the GRU/GRU-D encoder in the APC as the extracted representations, i.e. Enc($\textbf{x}) = \textbf{h}$, then implement the classifier on top of them. 
In the semi-supervised setting, we used two scenarios, consecutively:
\begin{enumerate}
\setlength\itemsep{0em}
    \item \textbf{frozen}: the weights of the pre-trained encoder are kept frozen and only the classification model is optimized.
    \item \textbf{fine-tuned}: APC's encoder and the classifier are trained end-to-end so that the learned representations are better suited to the final classification task. 
\end{enumerate}

\begin{figure}[t] 
\centering
\includegraphics[width=0.5\linewidth]{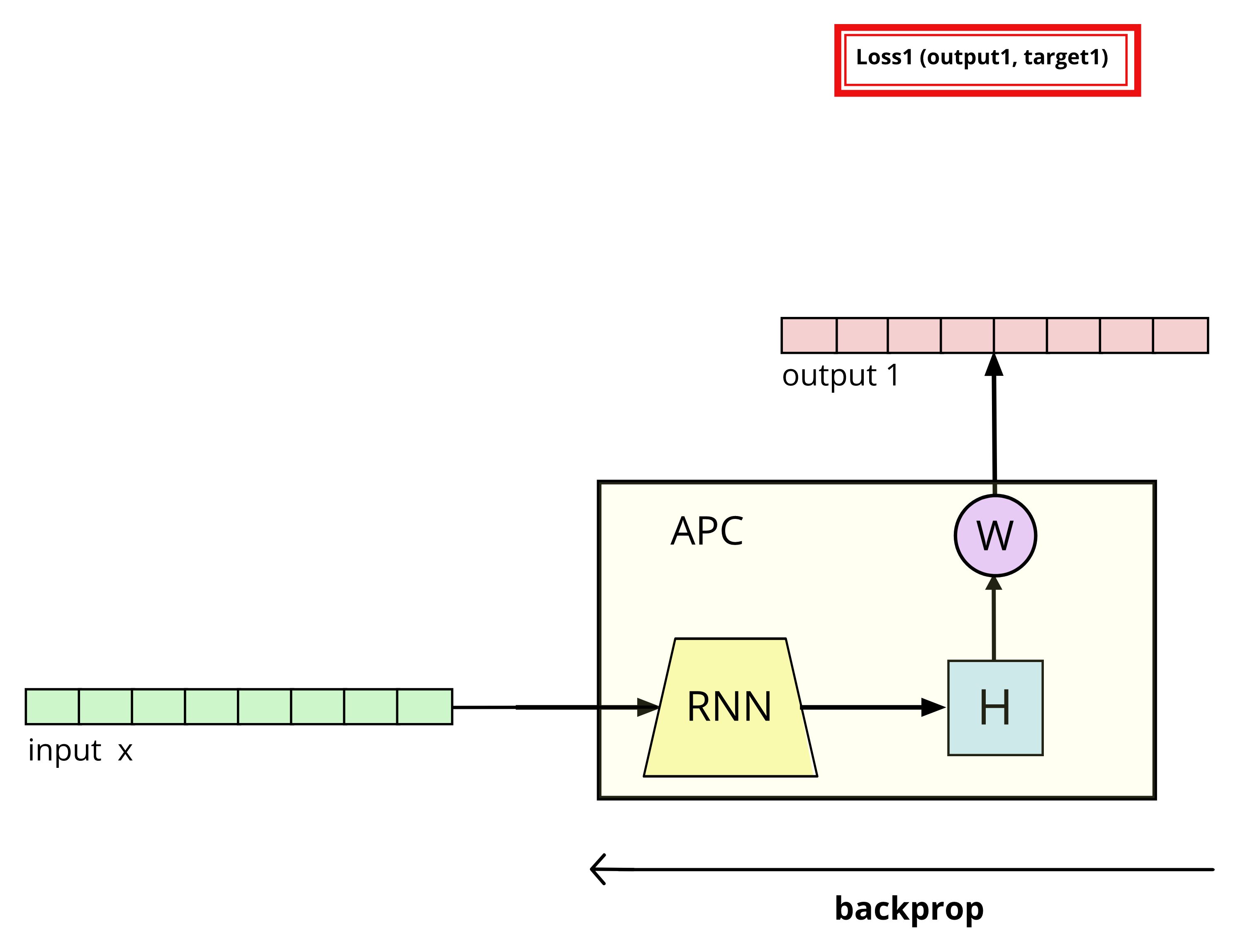}
\caption{\textbf{Step 1}: The APC is pre-trained by optimizing the autoregressive loss}
\label{apc_step_1}
\end{figure}

\begin{figure}[h] 
\centering
\includegraphics[width=0.6\linewidth]{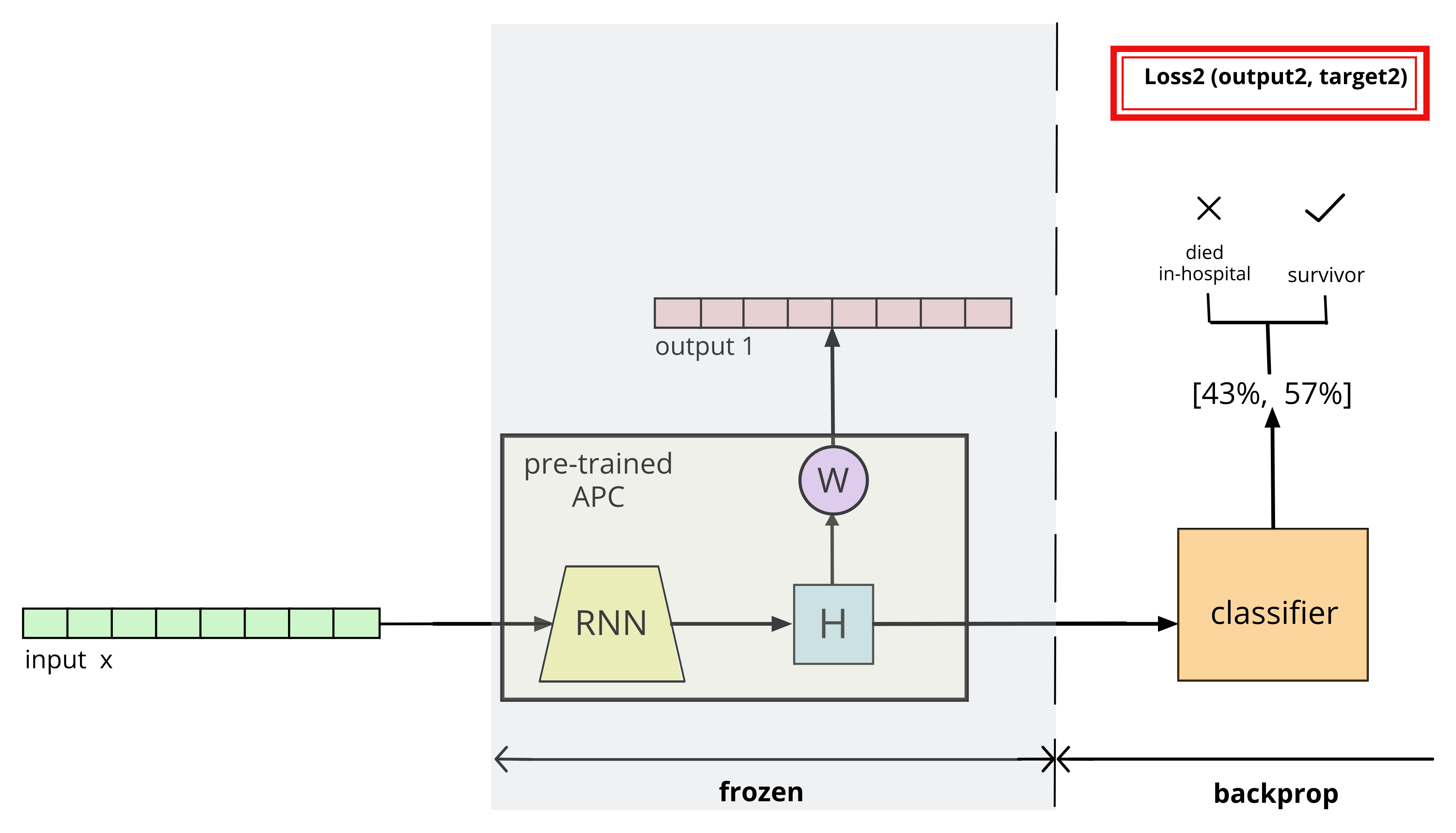}
\caption{\textbf{Step 2}: The APC encoder weights are frozen \& the classifier is trained by optimizing the cross-entropy loss.}
\label{apc_step_2}
\end{figure}

\begin{figure}[b] 
\centering
\includegraphics[width=0.6\linewidth]{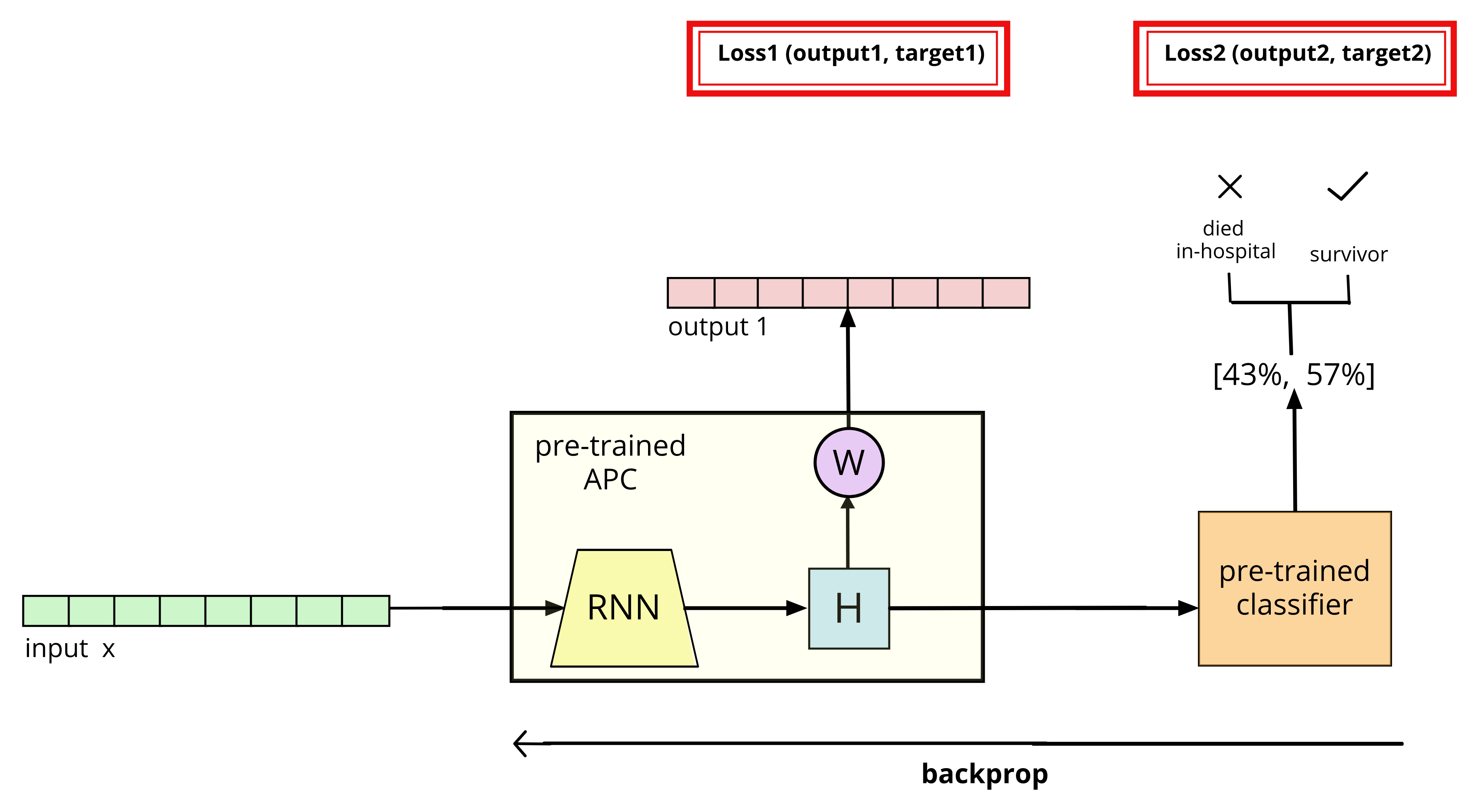}
\caption{\textbf{Step 3}: Both the encoder \& the classifier are trained end-to-end, using the pre-trained encoder \& pre-trained classifier weights.}
\label{apc_step_3}
\end{figure}

\clearpage
\subsection{\large Comparison to Referenced Publications}
\vspace{0.5cm}
To our knowledge, the most competitive and state-of-the-art results that have been achieved to date on the Physionet in-hospital mortality task are  by Horn et al. (2019), with their GRU-D and Transformer implementations attaining AUPRC scores of 53.7 $\pm$ 0.9, and 52.8 $\pm$ 2.2, respectively [12].  

Consequently, we set out to implement the GRU-D similar to their implementation, so that our methods are directly comparable.
However, we noticed that some architectural decisions taken in [12] were leading to lower performance scores overall compared to our initial implementation. 
Therefore, we decided to keep to our initial implementation. 
For full disclosure, the main differences between the two implementations are:
\begin{itemize}
    \item When processing the static variables, [12] compute the initial state of the RNN based on these static variables. Contrary to their implementation, our RNN does not handle the static variables. Instead, the static variables are concatenated with the final hidden states learned by the RNN which are then input directly into the classifier.
    \item To handle the class imbalance, they oversample the minority class, while we implement the GRU-D using class weights.
    \item Our dataset partition (train, validation, and test splits) might be different from theirs, which could be another source of potential variations in results.
\end{itemize}

\vspace{0.5cm}
\subsection{\large Evaluation Metrics}

\subsubsection*{\large Binary Classification}
The AUROC score is given by the following formula: 

\begin{equation} \text{AUROC} = \frac{1 + \text{TP}_\text{rate} - \text{FP}_\text{rate}}{2} = \frac{\text{TP}_\text{rate} + \text{TN}_\text{rate}}{2} 
\end{equation}

\noindent where the $\text{TP}_\text{rate}$, $\text{FP}_\text{rate}$, and the $\text{TN}_\text{rate}$ are defined as: 

\begin{equation}\label{recall}
\text{true positive rate (recall)} = \frac{\text{true positive}}{\text{true positive} + \text{false negative}} 
\end{equation} 

\begin{equation} \text{false positive rate} = \frac{\text{false positive}}{\text{false positive} + \text{true negative}} 
\end{equation}

\begin{equation} \text{true negative rate} = \frac{\text{true negative}}{\text{true negative} + \text{false positive}} 
\end{equation} 

\noindent However, when dealing with imbalanced classification, ROC curves may give an extremely optimistic view of the performance [2], and in these cases the precision-recall curves (PR curves) are recommended as the more appropriate metric. The PR curve shows a plot of the recall vs. precision for different probability tresholds. 
The recall, also known as the true positive rate or sensitivity, is displayed in Equation \ref{recall}, and the precision is defined as: 
\begin{equation}  
\text{precision} = \frac{\text{true positive}}{\text{true positive} + \text{false positive}}
\end{equation} 

\noindent Both the precision and recall are mostly focused on the positive class (minority class) and are not concerned about the true negatives (majority class). The Area Under this Precision-Cecall Curve score (AUPRC) is therefore mostly a representation of how well the model is handling the minority class. 
Since the AUPRC is arguably the most appropriate metric in the case of imbalanced classification, we focus on this performance metric for the binary classification task. 

\subsubsection*{\large Multiclass Classification}

We evaluate the performance of the multi-class classifier with the \textbf{F1-score} [20], which is the harmonic mean of the precision and recall:  

\begin{equation}
\textbf{F}_{1} = 2 \cdot \frac{\text{precision} \cdot \text{recall}}{ \text{precision} + \text{recall}}  
\end{equation}

\noindent Since our test set is class imbalanced, we do not want to give equal weights to each class, but instead we want to calculate a \textbf{weighted average F1-score}:  
\begin{equation}
\resizebox{0.7\hsize}{!}{$\frac{\#  \text{Class 1}}{\# \text{total}} \textbf{ x } \textbf{F1}_{\text{Class 1}} + \frac{\#  \text{Class 2}}{\# \text{total}} \textbf{ x } \textbf{F1}_{\text{Class 2}} + \frac{\#  \text{Class 3}}{\# \text{total}} \textbf{ x } \textbf{F1}_{\text{Class 3}} +  \frac{\#  \text{Class 4}}{\# \text{total}} \textbf{ x } \textbf{F1}_{\text{Class 4}}$}
\end{equation} 

\noindent Furthermore, to highlight the performances on the minority classes, we define the \textbf{weighted F1 minority}:
\begin{equation*}
\resizebox{0.7\hsize}{!}{$\frac{\#  \text{Class 2}}{\# \text{total minority}} \textbf{ x } \textbf{F1}_{\text{Class 2}} + \frac{\#  \text{Class 3}}{\# \text{total minority}} \textbf{ x } \textbf{F1}_{\text{Class 3}} +  \frac{\#  \text{Class 4}}{\# \text{total minority}} \textbf{ x } \textbf{F1}_{\text{Class 4}}$}
\end{equation*}  

\vspace{0.5cm}
\subsection{\large Data imputation baselines}
\vspace{0.4cm}
GRU-Mean: \begin{equation}
    x_{t}^{d} \xleftarrow{} m_{t}^{d} x_{t}^{d} + (1 - m_{t}^{d})\tilde{x}^{d}
\end{equation}

\noindent GRU-Forward: \begin{equation}
    x_{t}^{d} \xleftarrow{} m_{t}^{d} x_{t}^{d} + (1 - m_{t}^{d})x_{t'}^{d}
\end{equation}

\noindent GRU-Simple: \begin{equation}
    x_{t}^{(n)} \xleftarrow{} [x_{t}^{(n)}; m_{t}^{(n)}; \delta_{t}^{(n)}]
\end{equation}

\vspace{0.5cm}
\subsection{\large Oversampling \& class weights}
Because the Clue dataset contains a more severe class imbalance compared to Physionet, we propose another method specifically for Clue, which combines both oversampling [5] and using class weights. In this case, 
 we oversample the minority class to 12$\%$ and then use class weights on the oversampled class frequency.
 
\newpage
\vspace{0.5cm}
\subsection{\large Hyperparameters}
\vspace{1cm}
\begin{table*}[h]
\caption{\label{tab:chosen_hyperparameters} Final hyperparameters used for each model on each dataset, after performing hyperparameter optimization.}
\centering
\setlength{\tabcolsep}{10pt}
\begin{tabular}{lllll}
\toprule
{\textbf{Model}} & {Hyperparameters} & {Synthetic} & {Physionet} & {Clue} \\
\midrule
{\textbf{GRU}} & learning rate & 1e-03 & 1e-03 &  1e-04    \\
 & batch size  & 32 &   32  &   100\\
 & hidden units & 64 & 64 &   200\\
 & dropout & 0.0 &  0.0 &  0.4 \\
 & recurrent dropout & 0.0 &  0.0 & 0.1 \\
 & epochs & 150 &  50 & 200\\
\hline
{\textbf{GRU-D}} & learning rate & 1e-03 &  1e-03 & 1e-03 \\
& batch size  & 32 & 32 &  100 \\
& hidden units & 32 & 32 &  200\\
& dropout & 0.0 &  0.1 & 0.1 \\
& recurrent dropout & 0.0 &  0.0 &  0.0\\
& epochs & 100 &  50 & 200\\
\hline
{\textbf{GRU - APC}} & learning rate - step 1 & 1e-03 &  1e-03 &  1e-04 \\
& learning rate - step 2 $\&$ 3 & 1e-04 & 1e-04 &  1e-04   \\
& batch size & 32 &   32 &  100 \\
& hidden units & 120 &   64 &  200 \\
& dropout & 1.0 & 0.0,  & 0.4 \\
& recurrent dropout & 0.0 &  0.0 &  0.1\\
& epochs & 100 &  100 & 50\\
\hline
{\textbf{GRU-D - APC}} & learning rate - step 1 & 1e-03 &  1e-03 & 1e-03  \\
& learning rate - step 2 $\&$ 3 & 1e-04 & 1e-04 & 1e-04  \\
& batch size & 32 &  100 &  100  \\
& hidden units & 250 &  256  &  250 \\
& dropout & 0.0 & 0.1 &  0.1 \\
& recurrent dropout & 0.0 &  0.0 & 0.0 \\
& epochs - step 1 & 100 &  100 & 50\\
& epochs - step 2 $\&$ 3 & 100 &  50  &  50\\
\bottomrule 
\end{tabular}
\end{table*}

\newpage
\subsection{\large Results}

\begin{table}[h]
\caption{
Classification results on the synthetic dataset.}
\label{tab:results_synthetic}
\vskip 0.1in
\begin{center}
\begin{small}
\begin{sc}
\begin{tabular}{llllcc}
\toprule
{class} & {missing} &  & AUPRC (mean $\pm$ sd) &  &  \\
{(im)balance} & {data} &  &  &  &  \\
\hline
 & & GRU & GRU-D & GRU-APC & GRU-D-APC  \\
\midrule
{perfectly} & {0 $\%$} &  98.97 $\pm$ 0.1 &  99.01 $\pm$ 0.24 &  98.87 $\pm$ 0.39 &  98.85 $\pm$ 0.2\\ \cline{2-6}

{balanced} & {30 $\%$} &  98.4 $\pm$ 0.16 & 98.1 $\pm$ 0.22 & 98.24 
$\pm$ 0.53 &  98.07 $\pm$ 0.44 \\ \cline{2-6}
{(1:1)} & {60 $\%$} &  60.68 $\pm$ 12.6 & 90.33 $\pm$ 0.22 &  
96.67 $\pm$ 0.76 &  93.14 $\pm$ 1.67  \\ 
\hline
{slight} & {0 $\%$} &  97.32 $\pm$ 0.45 &  96.88 $\pm$ 0.61 & 97.26 $\pm$ 0.4 &  96.98 $\pm$ 1.21 \\ \cline{2-6}
{imbalance} & {30 $\%$} &  96.45 $\pm$ 0.58 & 92.22 $\pm$ 0.43 &   94.55 $\pm$ 1.33 &  91.03 $\pm$ 1.05 \\ \cline{2-6}
{(3:7)} & {60 $\%$} &  66.19 $\pm$ 10.54 &  80.55 $\pm$ 1.94 &   90.57 $\pm$ 0.52 &  83.06 $\pm$ 1.52\\ 
\hline
{severe} & {0 $\%$} &  76.05 $\pm$ 5.11 &  83.16 $\pm$ 2.03 & 92.31 $\pm$ 0.88 &  80.22 $\pm$ 2.95 \\ \cline{2-6}
{imbalance} & {30 $\%$} &  8.56 $\pm$ 0.58 &  53.05 $\pm$ 22.12 &  79.9 $\pm$ 3.8 &  63.94 $\pm$ 8.14 \\ \cline{2-6}
{(1:20)} & {60 $\%$} &  6.61 $\pm$ 0.5 &  35.54 $\pm$ 4.44 & 33.56 $\pm$ 21.06 &  66.42 $\pm$ 5.42\\
\bottomrule
\end{tabular}
\end{sc}
\end{small}
\end{center}
\vskip -0.15in
\end{table}

\vspace{0.5cm}

\begin{table}[h]
\caption{Comparison of using frozen versus fine-tuned APC weights on the Physionet classification results. 
These results also show the impact of using class weights as a class imbalance method.
All APC methods are implemented with time shift factor n = 1.}
\vskip 0.1in
\begin{center}
\begin{small}
\begin{sc}
\begin{tabular}{lllcc}
\toprule
model & APC weights method & class imbalance method & AUROC & AUPRC \\
\midrule
GRU & - & undersampling & 84.2 $\pm$ 0.3 & 48.4 $\pm$ 1.3\\ 
\hline 
GRU & - & oversampling & 84.6 $\pm$ 0.5 & 51.4 $\pm$ 1.7 \\
\hline 
GRU & - & class weights & 85.4 $\pm$ 0.4 & 51.4 $\pm$ 0.9\\ 
\hhline{=====}
GRU - APC  & frozen & none &  85.1 $\pm$ 0.2 &  51.8 $\pm$ 0.2\\ 
\hline 
GRU - APC & frozen & class weights &  85.0 $\pm$ 0.2 & 51.3 $\pm$ 0.1 \\
\hline 
GRU - APC  & fine-tuned & none & \textbf{86.0} $\pm$ \textbf{0.5} & \textbf{54.1} $\pm$ \textbf{1.0}\\ 
\hline 
GRU - APC & fine-tuned & class weights & 85.9 $\pm$ 0.3 & 53.5 $\pm$ 0.5\\ 
\hhline{=====}
GRU-D - APC & frozen & none & 85.1 $\pm$ 0.9 & 54.0 $\pm$ 2.3\\ 
\hline 
GRU-D - APC & frozen & class weights &  85.3 $\pm$ 0.1 & \textbf{55.1} $\pm$ \textbf{0.9}\\ 
\hline 
GRU-D - APC & fine-tuned & none & 85.2 $\pm$ 0.9  & 54.1 $\pm$ 2.3\\ 
\hline 
GRU-D - APC & fine-tuned & class weights & 85.3 $\pm$ 0.2 & 55.0 $\pm$ 0.9\\
\bottomrule
\end{tabular}
\end{sc}
\end{small}
\end{center}
\vskip -0.15in
\end{table}

\begin{table}[h]
\caption{
Classification results on the Clue dataset using \textbf{GRU} - \textbf{APC} for different values of the time shift factor.}
\label{tab:results_clue_APC_GRU}
\vskip 0.1in
\begin{center}
\begin{small}
\begin{sc}
\begin{tabular}{llllcccc}
\toprule
{APC weights} & {time} &  {F1 ON} & {F1 OFF} & {F1} & {F1} & {weighted} & {weighted}\\
{Method} & {shift} &  &  & {OTHER H} & {OTHER NH} & {F1} & {F1 minority}\\
\midrule
{frozen} & 0 & 96.7 $\pm$ 0.0 & 19.7 $\pm$  4.6 &  28.2 $\pm$ 1.7 & 17.1 $\pm$ 1.4 & 90.4 $\pm$ 0.2 & 21.5 $\pm$ 2.1 \\
\hline
{fine-tuned} &  0 & 96.7 $\pm$ 0.1 & 22.0 $\pm$ 12.3 & 29.8 $\pm$ 2.4 & 14.6 $\pm$ 0.4 & 90.4 $\pm$ 0.7 & 22.8 $\pm$ 6.6 \\
\hline
{frozen} & 1 & 96.7 $\pm$ 0.0 & 24.8 $\pm$ 6.4 &  24.1 $\pm$ 3.2 & 16.6 $\pm$ 2.9 & \textbf{90.5} $\pm$ \textbf{0.3} & \textbf{23.2} $\pm$ \textbf{3.0}\\
\hline
{fine-tuned} &  1 & 96.7 $\pm$ 0.0 & 28.8 $\pm$ 2.2 & 25.7 $\pm$ 2.4 & 14.8 $\pm$ 2.5 & \textbf{90.7} $\pm$ \textbf{0.1} & \textbf{25.7} $\pm$ \textbf{1.1} \\
\hline
{frozen} &  2 & 96.7 $\pm$ 0.1 & 27.3 $\pm$ 4.0 &  22.9 $\pm$ 3.3 & 16.2 $\pm$ 3.5 & 90.6 $\pm$ 0.3 & 24.3 $\pm$ 2.5 \\
\hline
{fine-tuned} &  2 & 96.6 $\pm$ 0.3 & 23.3 $\pm$ 14.0 &  24.0 $\pm$ 2.4 & 14.6 $\pm$ 1.5 & 90.3 $\pm$ 0.9 & 22.1 $\pm$ 7.3\\
\hline
{frozen} &  5 & 96.5 $\pm$ 0.2 & 24.7 $\pm$ 5.9 & 17.4 $\pm$ 9.7 & 17.4 $\pm$ 2.2 & 90.2 $\pm$ 0.5 & 21.6 $\pm$ 4.5 \\
\hline
{fine-tuned} &  5 & 96.0 $\pm$ 0.7 & 16.6 $\pm$ 8.9 & 14.8 $\pm$ 13.0 & 15.0 $\pm$ 0.6 & 89.2 $\pm$ 1.3 & 15.9 $\pm$ 8.2\\
\bottomrule
\end{tabular}
\end{sc}
\end{small}
\end{center}
\vskip -0.15in
\end{table}

\begin{table}[h]
\caption{
Classification results on the Clue dataset using \textbf{GRU-D} - \textbf{APC} for different values of the time shift factor.}
\label{tab:results_clue_APC_GRUD}
\vskip 0.1in
\begin{center}
\begin{small}
\begin{sc}
\begin{tabular}{llllcccc}
\toprule
{APC weights} & {time} &  {F1 ON} & {F1 OFF} & {F1} & {F1} & {weighted} & {weighted}\\
{Method} & {shift} &  &  & {OTHER H} & {OTHER NH} & {F1} & {F1 minority}\\
\midrule
{frozen} & 0 & 95.6 $\pm$ 0.6 & 28.1 $\pm$ 2.3 & 20.1 $\pm$ 2.6 & 11.6 $\pm$  2.4 & 89.5 $\pm$ 0.7 & 23.3 $\pm$ 2.4\\
\hline
{fine-tuned} &  0 & 96.2 $\pm$ 0.1 & 32.4 $\pm$ 0.8 & 23.1 $\pm$ 2.1 & 14.4 $\pm$ 2.6 & 90.3 $\pm$ 0.1 & 27.0 $\pm$ 0.5\\
\hline
{frozen} & 1 &  96.2 $\pm$ 0.0 & 32.0 $\pm$ 0.0 & 25.4 $\pm$ 0.2 & 13.9 $\pm$  2.8 &  \textbf{90.3} $\pm$ \textbf{0.0} & \textbf{27.3} $\pm$ \textbf{0.5} \\
\hline
{fine-tuned} &  1 & 96.0 $\pm$ 0.5 & 30.3 $\pm$ 0.2 & 21.7 $\pm$ 4.0 & 13.7 $\pm$ 2.2 & \textbf{90.0} $\pm$ \textbf{0.5} & \textbf{25.3} $\pm$ \textbf{0.8} \\
\hline
{frozen} &  2 &  96.1 $\pm$ 0.1 & 31.4 $\pm$ 0.1 & 24.8 $\pm$ 0.5 & 10.7 $\pm$  0.8 & 90.2 $\pm$ 0.0 & 26.3 $\pm$ 0.2\\
\hline
{fine-tuned} &  2 &  95.7 $\pm$ 0.2  & 31.2 $\pm$ 0.6  &  19.2 $\pm$ 1.0 & 11.4 $\pm$ 2.6  & 89.7 $\pm$ 0.3 & 24.9 $\pm$ 1.1 \\
\hline
{frozen} &  5 & 95.7 $\pm$ 0.4 & 31.6 $\pm$ 1.2 & 18.6 $\pm$ 4.8 & 11.2 $\pm$ 3.7 &  89.7 $\pm$ 0.5 & 24.8 $\pm$ 1.1\\
\hline
{fine-tuned} &  5 & 95.8 $\pm$ 0.1 & 32.6 $\pm$ 0.2 & 17.0 $\pm$ 1.0 & 10.8 $\pm$ 1.5 & 89.8 $\pm$ 0.2 & 25.0 $\pm$ 0.6\\
\bottomrule
\end{tabular}
\end{sc}
\end{small}
\end{center}
\vskip -0.15in
\end{table}

\begin{figure}[h]
\begin{center}
\centerline{\includegraphics[width=0.9\textwidth]{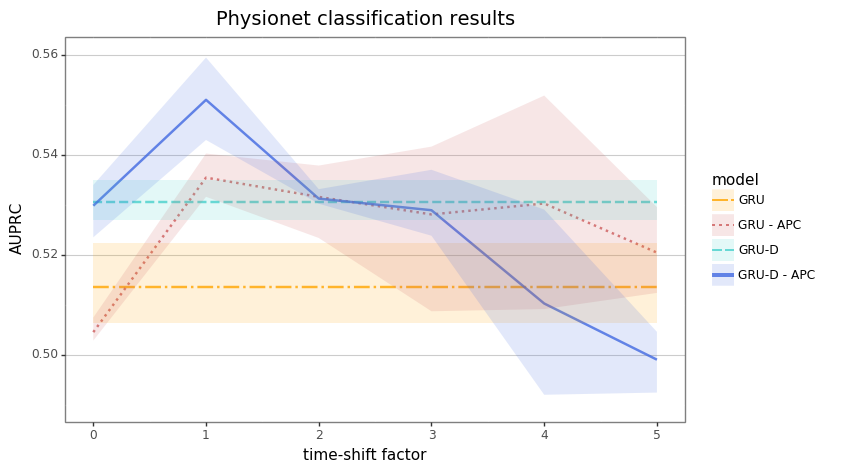}}
\vspace{-4mm}
\caption{Physionet classification results for different values of the time shift factor \textit{n}. All models are implemented with class weights.} 
\vspace{-2.5em}
\label{physionet_APC_plot}
\end{center}
\end{figure}

\end{document}